\pdfoutput=1

\documentclass[11pt]{article}

\usepackage{coling}

\usepackage{times}
\usepackage{latexsym}

\usepackage[T1]{fontenc}

\usepackage[utf8]{inputenc}
\usepackage{times}
\usepackage{latexsym}
\usepackage{url}
\usepackage{enumitem}
\usepackage{amsmath,amssymb,amsfonts}
\usepackage{algorithmic}
\usepackage{caption}
\usepackage{subcaption}
\usepackage{graphicx}
\usepackage{textcomp}
\usepackage{url}
\usepackage{makecell}
\usepackage{multirow}
\usepackage{array}
\usepackage{microtype}
\usepackage{colortbl}
\usepackage{multirow}
\usepackage{booktabs}

\usepackage{listings}       

\usepackage{hyperref}       
\usepackage{amsfonts}       
\usepackage{nicefrac}       
\usepackage{xcolor}         
\usepackage{tcolorbox}

\newcommand{\thickhline}{%
    \noalign {\ifnum 0=`}\fi \hrule height 1pt
    \futurelet \reserved@a \@xhline
}

\newcommand{\inlinecode}[2]{\colorbox{lightgray}{\lstinline[language=#1]$#2$}}

\usepackage{pifont}
\newcommand{\cmark}{O}%
\newcommand{\xmark}{\ding{55}}%
\definecolor{ForestGreen}{RGB}{34,139,34}

\newcommand\blfootnote[1]{%
  \begingroup
  \renewcommand\thefootnote{}\footnote{#1}%
  \addtocounter{footnote}{-1}%
  \endgroup
}
\usepackage{listings}
\definecolor{codegreen}{rgb}{0,0.6,0}
\definecolor{codegray}{rgb}{0.5,0.5,0.5}
\definecolor{codepurple}{rgb}{0.58,0,0.82}
\definecolor{backcolour}{rgb}{0.95,0.95,0.92}

\lstdefinestyle{mystyle}{
    backgroundcolor=\color{backcolour},   
    commentstyle=\color{codegreen},
    keywordstyle=\color{magenta},
    numberstyle=\tiny\color{codegray},
    stringstyle=\color{codepurple},
    basicstyle=\ttfamily\footnotesize,
    breakatwhitespace=false,         
    breaklines=true,                 
    captionpos=b,                    
    keepspaces=true,                 
    numbers=left,                    
    numbersep=5pt,                  
    showspaces=false,                
    showstringspaces=false,
    showtabs=false,                  
    tabsize=2
}

\usepackage{booktabs}
\usepackage{graphicx}
\usepackage{pifont}

\definecolor{lp}{HTML}{CBC3E3}
\usepackage{colortbl}
\usepackage{pythonhighlight}
\usepackage{xspace}

\usepackage{amsmath}
\usepackage{amsfonts}
\definecolor{lp}{HTML}{CBC3E3}

\title{Dataverse: Open-Source ETL (Extract, Transform, Load) Pipeline for Large Language Models}

\author{Hyunbyung Park$^{1}$, Sukyung Lee$^{2}$, Gyoungjin Gim$^{2}$ \\ {\bf \large Yungi Kim$^{3}$, Dahyun Kim$^{4}$, Chanjun Park$^{5\dagger}$}\\
\\
  $^{1}$Moreh, $^{2}$Upstage AI, $^{3}$Liner, $^{4}$Twelve Labs, $^{5}$Korea University \\
  \texttt{\normalsize{hyunbyung.park@moreh.io}}, \texttt{\normalsize\{sukyung, gyoungjin.gim\}@upstage.ai} \\ \texttt{\normalsize{eddie}@linercorp.com}, \texttt{\normalsize{kian.kim}@twelvelabs.io} \\ 
  \texttt{\normalsize{bcj1210}@korea.ac.kr}
  }

\begin{document}
\maketitle
\begin{abstract}
\blfootnote{$^\dagger$ Corresponding Author}
To address the challenges associated with data processing at scale, we propose Dataverse\footnote{\url{https://github.com/UpstageAI/dataverse}}, a unified open-source Extract-Transform-Load (ETL) pipeline for large language models (LLMs) with \textit{a user-friendly design} at its core. Easy addition of custom processors with block-based interface in Dataverse allows users to readily and efficiently use Dataverse to build their own ETL pipeline. We hope that Dataverse will serve as a vital tool for LLM development and open source the entire library to welcome community contribution. Additionally, we provide a concise, two-minute video demonstration of our system, illustrating its capabilities and implementation\footnote{\url{https://www.youtube.com/watch?v=yYyyLuPNK5s&t=33s}}.

\end{abstract}

\section{Introduction}
The success of large language models (LLMs) is widely attributed to the scale of the data~\cite{zhao2023survey}, otherwise known as the \textit{`scaling law'}~\cite{kaplan2020scaling} where LLM performance directly correlates with data size. Consequently, there has been an exponential growth in the need for massive data to further fuel LLM development. Such increase in demand leads to more complex data processing pipelines, as even simple operations need to be optimized for data processing at enormous scales. To handle such data workloads efficiently and effectively, distributed systems and techniques such as Spark~\cite{zaharia2016apache} and Slurm~\cite{yoo2003slurm} have become crucial.

Unfortunately, the existing open-source data processing tools based on distributed systems~\cite{chenghao_mou_2023_8364980, dolma, lee2021deduplicating, penedo2024datatrove} either lack easy customization support or a wide variety of operations such as deduplication~\cite{xia2016comprehensive}, decontamination~\cite{yang2023rethinking}, bias mitigation~\cite{shrestha2022investigation}, and toxicity reduction~\cite{wang2022toxicity}. This forces researchers to undergo a steep learning curve or cobble together tools from various sources, hindering efficiency and user experience.

In response to these limitations, we present Dataverse, a \textit{unified} open-source ETL (Extract, Transform, Load) pipeline with \textit{a user-friendly design} that enables easy customization. 
Inspired by the Transformers library~\cite{wolf2019huggingface}, Dataverse is built with a design principle of minimizing complex inheritance structures. Such design choice allows for easy addition of custom data operations. Specifically, the  ETL pipeline in Dataverse is defined by block-based interface, which enables intuitive customization of ETL pipelines by simply adding, removing, or reshuffling blocks. Further, Dataverse natively supports a wide range operations needed to cover diverse data processing use-cases. 

Moreover, the data processing workloads can be distributed among multiple nodes with Spark by simply setting the necessary configurations.
Further, user-friendly debugging features via Jupyter notebooks are included for fast build-test of custom ETL pipelines. In addition, Dataverse supports multi-source data ingestion from on-premise storage, cloud platforms, and even web scraping. This feature empowers users to easily transform raw data from various sources. 
Driven by the aforementioned features, we posit that Dataverse will be a useful tool for effortlessly building custom ETL pipelines at scale for fast LLM development.

\begin{table*}[t!]
\centering
\resizebox{0.7\linewidth}{!}{
    \centering
    \begin{tabular}{cccc}
    \toprule 
         Open-Source Library & Dist. System & Expandable & Customization Difficulty \\ \midrule
         text-dedup & Spark & \color{red} \xmark & N/A\\
         DPS & Spark  & \color{red} \xmark & N/A \\
         deduplication-text-datasets & Rust & \color{red} \xmark & N/A\\
         Dolma & Rust & \color{red} \xmark & N/A\\
         Datatrove & Slurm  & \color{ForestGreen} \cmark & High\\
         \cellcolor{lp!60}Dataverse  & \cellcolor{lp!60}Spark & \cellcolor{lp!60}{\color{ForestGreen} \cmark} &\cellcolor{lp!60} Low\\
    \bottomrule
    \end{tabular}
}
\caption{Comparison between existing open-source LLM data processing libraries and Dataverse. ``Dist. System'', ``Expandable'', ``Customization Difficulty'' indicate the distributed system integrated into the library, whether the library is designed to be future-proof and capable of growth, and the difficulty of the customization, respectively. N/A means customization is not natively supported.}
\label{tab:data_libraries_open_source}
\end{table*}

\section{Why Dataverse?}
In the era of LLMs, data scales exponentially~\cite{kaplan2020scaling}, necessitating an efficient and scalable solution~\cite{wang2023data}. Not only that, the fast pace of the LLM literature comes with the need to support a wide range of data operations such as toxicity and bias removal~\cite{garg2023handling}, personally identifiable information (PII) obfuscation~\cite{schwartz2011pii}, and data quality filterings~\cite{shin2022effect,choi2023dmops}.
Thus, on top of utilizing distributed systems, LLM-aware data processing also requires natively supporting a wide variety of operations and easy addition of custom data operations.

While there are many existing data processing libraries such as proposed~\cite{chenghao_mou_2023_8364980, dolma, lee2021deduplicating, penedo2024datatrove}, none of them are not yet the whole package of being easy to customize and supporting a wide variety data operations. To address this gap, we introduce Dataverse with a user-friendly design in mind, allowing users to utilize (custom) data processing tools and distributed system via simply setting the blocks and configurations.
In the following sections, we detail the comparison between Dataverse and other existing open-source frameworks for LLM-aware data processing.

\begin{figure*}[t]
    \centering
    \resizebox{0.79\linewidth}{!}{
        \includegraphics{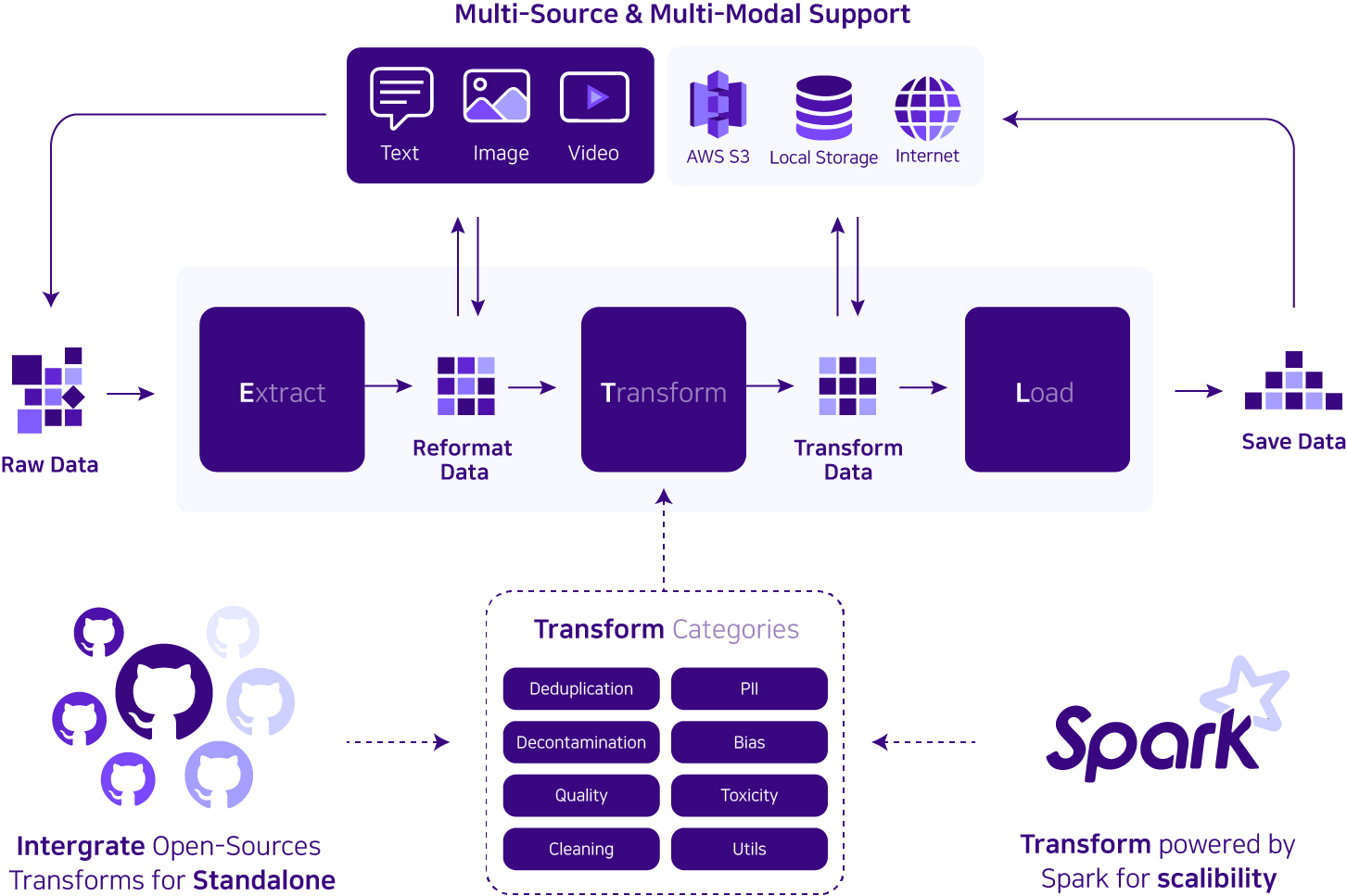}
    }
    \caption{Overview of the Dataverse library.}
    \label{fig:dataverse_overview}
\end{figure*}

\subsection{Comparison Between Dataverse and  Other Open Source Libraries}
As explained in the previous section, data processing libraries for the LLMs need to support a wide variety of data operations and distributed systems for scalable data processing.
Further, the library itself needs to be expandable to accommodate novel data processing operations as they emerge.
Lastly, taking one step further from just being expandable, it would be ideal if such expansion to custom data processing operations can be made effortlessly.
We compare various open source data processing libraries and Dataverse in the aforementioned criteria in Table~\ref{tab:data_libraries_open_source}.

As shown in the table, the compared open source libraries, including text-dedup~\cite{chenghao_mou_2023_8364980}, DPS\footnote{\url{https://github.com/EleutherAI/dps}}, deduplicate-text-datasets~\cite{lee2021deduplicating}, Dolma~\cite{dolma}, and Datatrove~\cite{penedo2024datatrove}, as well as Dataverse all support distributed system.
Specifically, text-dedup, DPS, and Dataverse use Spark~\cite{zaharia2016apache} as the distributed system of choice, while deduplication-text-datasets utilizes parallel processing of Rust~\cite{matsakis2014rust} and Datatrove uses Slurm~\cite{yoo2003slurm} for their distributed system.

The comparison becomes clearer once we look at the ``expandable'' criteria. \textbf{Expandable} means rather than being a static library provided \textbf{as is}, rather dynamic, evolving library inherently designed to grow and adapt over time. Specifically, libraries such as text-dedup, deduplication-text-datasets, and Dolma all lack expandability as they are developed for one-time use, for instance, academic purposes. In contrast, Datatrove and Dataverse both support expanding the library suitable for future-proof LLM data handling. They feature interfaces that facilitate ongoing modification of the library. Also, they encourage community engagement by providing guidelines and processes for contribution ensuring the library remains adaptable and up-to-date.

Further, we compare how difficult it is to customize the library for a user's data processing workload.
Note that for libraries that are not expandable, comparing the customization difficulty is not applicable.
The customization difficulty for Datatrove is high as a user needs to make code changes to multiple places while adhering to the complex inheritance design of the Datatrove library.
Conversely, the customization difficulty for Dataverse is low as the user simply needs to define a custom data processing operation function and register it to the Dataverse library using a decorator, as illustrated in Section~\ref{sec:add_custom_op}.
We now explain the key features and system architecture of Dataverse in the following sections.

\section{Dataverse}\label{sec:section3}
Dataverse is an open-source library for building ETL pipeline for LLMs with user-friendly design at its core. The overview of the Dataverse library is shown in Figure~\ref{fig:dataverse_overview}.

\subsection{Key Features}
\paragraph{User-friendly design.}
The user-friendly design of Dataverse is implemented in consideration to various aspects.
First, various tools necessary for building a complete ETL pipeline are optimized and unified such that users can use Dataverse as a standalone solution for building their custom ETL pipelines.
As such, Dataverse natively supports optimized functions for various steps in the data processing workflow such as data downloading, reformatting, processing, and storing, ridding the need to look for other solutions even at very large data scales.
Detailed explanation on the supported functionalities is given in Section~\ref{sec:supported_op}.

Second, to support easy customization of ETL pipelines, Dataverse incorporates a strikingly simple method of adding custom data processing functions via Python decorators.
Thus, users can readily utilize custom functions beyond that of the already large number of natively supported operations that are registered.

Third, utilizing either the natively supported operation or an custom added function to create an ETL pipeline in Dataverse is intuitive and flexible. The reason is that ETL pipelines in Dataverse are implemented using a block-based interface such that users can define a modular \textit{block}, an atomic unit of data processing.
Then, users can change their ETL pipeline by re-organizing the defined blocks, allowing for straightforward development of data processing pipelines.
Further, Dataverse supports local testing functionality via Jupyter notebooks which allows users to inspect their ETL pipeline at various stages before scaling out.

\begin{figure*}[t]
    \centering
    \resizebox{0.85\linewidth}{!}{
        \includegraphics{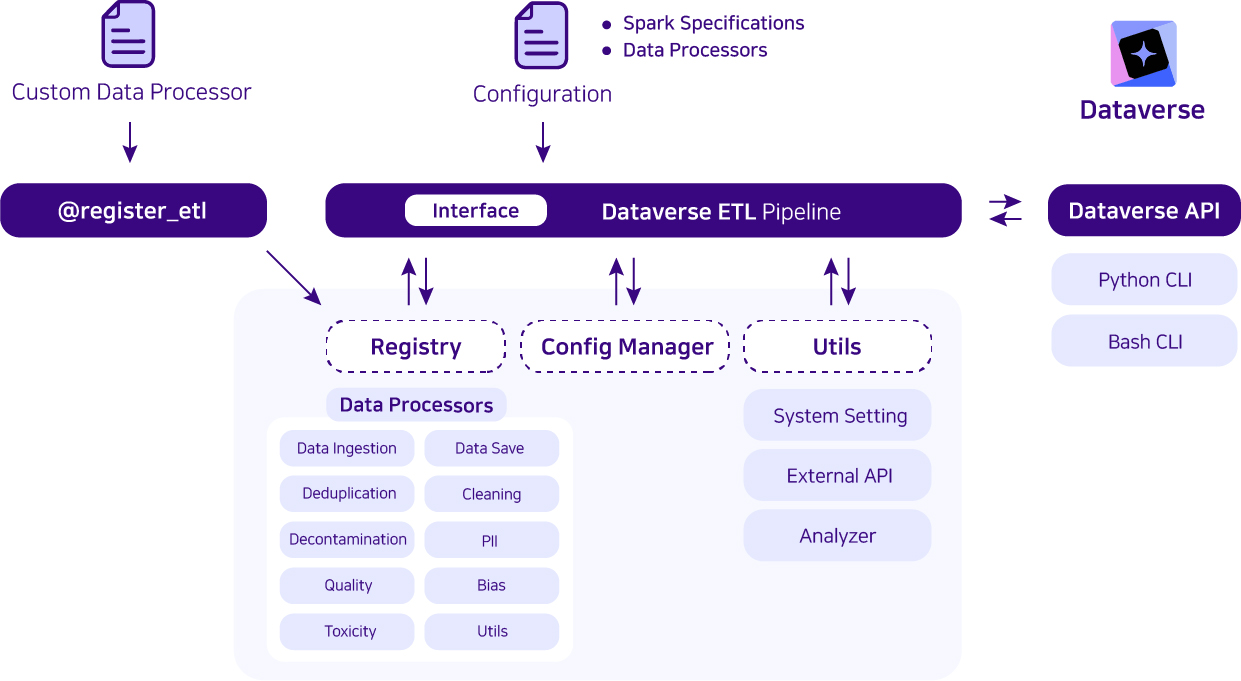}
    }
    \caption{Architecture of Dataverse.}
    \label{fig:dataverse_architecture}
\end{figure*}

\paragraph{Scalability via Spark and AWS Integration}
To scale ETL pipelines efficiently, Dataverse leverages Apache Spark~\cite{zaharia2016apache}, enabling distributed processing capabilities. Furthermore, it natively integrates with Amazon Web Services (AWS) for cloud utilization, facilitating greater scalability. Currently, Dataverse supports AWS S3 for cloud storage and Elastic MapReduce (EMR) for data processing. This integration ensures that users without access to sufficient local computing resources can effectively manage their data without encountering steep limitations.
The aforementioned features can be enabled by simply changing the configuration or giving an argument when running the ETL pipeline.

\subsection{System Architecture}
Figure \ref{fig:dataverse_architecture} illustrates the overall system architecture of Dataverse.

\paragraph{ETL pipeline.}
The ETL pipeline represents the primary interface for Dataverse users. This central core interface facilitates communication with various modules, including configuration, registry, application programming interface (API), and utilities. Its primary objective is to ensure the seamless creation and operation of the ETL pipeline, effectively managing data processing tasks. Additionally, the interface offers AWS EMR integration by simply passing the ``True'' value to the ``emr'' option, as described in Section~\ref{sec:library_tour}. This straightforward approach empowers users to leverage the scalability of cloud computing without the steep learning curve typically associated with distributed systems management.

\paragraph{Configuration.}
The user prepares a configuration object that encapsulates all the essential details required to execute the ETL Pipeline. The configuration facilitates the setup of Apache Spark specifications and the selection of the data processors to be employed.

\paragraph{Configuration manager.}
The configuration manager manages various configurations from specified paths (local, AWS S3) or handles multiple types (Python Dict, YAML, OmegaConf) of configuration data. It converts these configurations into a unified format compatible with Dataverse, ensuring they are ready for use in the system.

\paragraph{Registry.}
\label{sec:supported_op}
The registry serves as a repository where all data processor functions are stored.
The data processors to be utilized are specified within the configuration which are then retrieved from the registry to assemble the desired ETL pipeline.
Note that custom data processors can be added by simply using the \inlinecode{python}{@register_etl} decorator.
The list of natively supported data processors is as follows:
\begin{itemize}
    \item \textbf{Data Ingestion}: Facilitating the loading of data from various sources (\textit{e.g.}, data in Huggingface Hub, and parquet/csv/arrow format data in local storage) into a preferred format.
    \item \textbf{Data Saving}: Persisting the processed data into a preferred destination, such as a data lake or database.
    \item \textbf{Deduplication}: Eliminating duplicated data on dataset by dataset basis or globally across multiple datasets.
    \item \textbf{Data Cleaning}: Removing irrelevant, redundant, or noisy information from the data, such as stop words or special characters.
    \item \textbf{Data Decontamination}: Identifying and removing contaminated data such as benchmark datasets.
    \item \textbf{Personally Identifiable Information (PII) Removal}: Ensuring the removal of sensitive information, such as personally identifiable data, from the dataset.
    \item \textbf{Data Quality Enhancement}: Improving the quality of data from the perspectives of accuracy, consistency, and reliability for LLMs.
    \item \textbf{Bias Mitigation}: Reducing skewed or prejudiced data, with a particular emphasis on data that reinforces stereotypes of LLMs.
    \item \textbf{Toxicity Removal}: Identifying and eliminating harmful, offensive, or inappropriate content within the data.
    \item \textbf{Utilities}: Providing essential functionalities for data processing, including sampling, logging, and statistical analysis.
\end{itemize}

\paragraph{Utilities.}
The Utilities module serves as an internal helper toolset. One of its core features is the API utilities, which streamline the use of various external APIs such as AWS EMR. It simplifies the deployment and management of AWS EMR, reducing the complexity for researchers unfamiliar with cloud infrastructure. By simply setting their own AWS Credentials, Dataverse automatically handles the intricate details of provisioning EMR clusters and orchestrating the data processing pipelines across the cluster nodes.

\paragraph{Dataverse API.}
The Dataverse API serves as a gateway for users. Currently, Dataverse supports the Python CLI (Command Line Interface). Additionally, the development of a Bash CLI is underway.

\subsection{Library Tour}\label{sec:library_tour}
The Dataverse interface is designed to be intuitive and user-friendly, substantially simplifying the data processing workflow. Careful consideration has been given to user experience, minimizing the learning curve for new users and enabling them to rapidly understand and effectively utilize Dataverse with minimal effort.

\paragraph{Executing ETL pipeline with configuration.}
Using Dataverse is straightforward, primarily requiring a properly designed configuration for the execution of the ETL pipeline. The essential configuration elements include specifying the Apache Spark specifications for execution and ordering the data processors to be applied. The initial data processor must be configured for data ingestion to facilitate data loading, followed by any additional data processors the user wishes to employ.
We give an example of using Dataverse below, with the configuration simplified for brevity.

\begin{lstlisting}[language=bash,numbers=none]
# import necessary libraries
import OmegaConf
from dataverse.etl import ETLPipeline

# set up configuration
config = OmegaConf.create({
    'spark': {Spark spec},
    'etl': [
        {data ingestion}
        {cleaning}
        {deduplication}
        {data saving}
      ]
  })

# run on ETL pipeline
etl = ETLPipeline()
spark, dataset = etl.run(config)
\end{lstlisting}

\paragraph{Incorporating custom data processors.}
\label{sec:add_custom_op}
Integrating a custom data processor into Dataverse requires defining a custom function and decorating it using \inlinecode{python}{@register_etl}. The custom function requires only two mandatory inputs, a Spark instance and the input data. 
Thus, creating custom operations in Dataverse is a natural extension for those with proficiency in Spark.
An example of adding custom processors is given below.

\begin{lstlisting}[language=bash,numbers=none]
# add your custom process
@register_etl
def add___one___func(spark, x):
    x = x.map(lambda x: x + 1)
    
    return x

# add to configuration
config = OmegaConf.create({
    'spark': {Spark spec},
    'etl': [
        {data ingestion}
        {add___one___func}
        {cleaning}
        {deduplication}
        {data saving}
      ]
  })

# run on ETL pipeline
etl = ETLPipeline()
spark, dataset = etl.run(config)
\end{lstlisting}

\paragraph{Scaling with AWS EMR.}
As explained in the previous section, Dataverse natively supports AWS integration to provide a solution for users facing local resource limitations. To leverage the power of AWS EMR, users can simply add a single argument when running their own ETL pipeline. An example usage is given below.

\begin{lstlisting}[language=bash,,numbers=none]
# run on AWS EMR
etl = ETLPipeline()
etl.run(config, emr=True)
\end{lstlisting}

\paragraph{Debugging with helper functions.}
To facilitate debugging, Dataverse provides helper functions such as generating fake data. Further, users can start debugging at any point within the pipeline by retaining only the steps up to the point they wish to debug in their own ETL pipeline.

\begin{lstlisting}[language=bash,numbers=none]
config = OmegaConf.create({
    'spark': {Spark spec},
    'etl': [
        {generate_fake_data} 
      ]
  })
etl = ETLPipeline()
spark, x = etl.run(config, emr=True)

# start debugging with your output from this line
print(x.show())
\end{lstlisting}

\section{Related Work and Background}\label{sec:related_work}
\subsection{Distributed Processing for Massive Datasets}
The processing of big data has presented significant challenges since the advent of the internet era. In the early stages of deep learning, models were developed for specific purposes using relatively small datasets. However, the emergence of large language models (LLMs) has necessitated the use of massive datasets, rendering distributed processing an indispensable requirement. Rather than relying on single nodes, multi-node and multi-processing environments enabled by open-source tools such as Slurm~\cite{yoo2003slurm} and Spark~\cite{zaharia2016apache} have become essential. LLM-aware data processing tools have been designed with distributed processing architectures in mind to address the immense computational demands.

\subsection{Data Quality Control for Large Language Models}
Ensuring data quality at a massive scale presents formidable challenges. Manual inspection of the data is impractical due to its sheer volume. The emphasis on data quality control has become crucial~\cite{penedo2023refinedweb,choi2023dmops}, primarily because the pursuit of larger datasets often involves incorporating low-quality data that has not undergone meticulous human curation~\cite{li2024quantity,chung2022scaling}. One of the most notable examples of such massive datasets is Common Crawl~\cite{dodge2021documenting}, often regarded as the holy grail of web data. However, this indiscriminately crawled data from the internet frequently suffers from a myriad of issues, including duplicated content, excessive brevity or verbosity, hidden biases, and the inclusion of junk data. To address these challenges, implementing a range of strategies for data quality enhancement is essential, among which deduplication is particularly critical~\cite{lee2022deduplicating}. Even when utilizing high-quality datasets, the possibility of encountering duplicated data remains, as multiple sources may be incorporated. Another key strategy could be the elimination of benchmarks or other unintended data inadvertently included in the dataset, which is known as decontamination. Additionally, removing overly short or excessively long sentences could be essential for maintaining data integrity~\cite{MOON2023120962,li2024superfiltering}.

\subsection{ETL (Extract, Transform, Load)}
ETL, stands for Extract, Transform, Load, is a fundamental process that involves gathering data from various sources and consolidating it. In the step of ``Extract'', Dataverse retrieves raw data and prepares it for processing. During ``Transform'', the data undergoes various processes such as deduplication and cleaning. Finally, Dataverse performs ``Load'' step which transfers the processed data into a storage destination of choice. Incorporating all these ETL steps enables end-to-end data handling from multi-sources.

\section{Conclusion}
To address the surging needs of data processing at massive scales, owing to the rise in popularity of LLMs, we propose Dataverse, an open-source library for ETL pipelines designed with scalability and future growth in mind. Dataverse is user-friendly designed with a block-based interface, allowing users to easily add custom functions for data processing while also natively supporting a wide array of commonly used data operations. Furthermore, Dataverse offers scalable solutions through its seamless integration with Spark and AWS EMR, allowing users to use Dataverse to handle data workloads of varying sizes. We envision Dataverse becoming a central hub for LLM data processing, facilitating collaboration, knowledge exchange, and ultimately accelerating advancements in the field.

\section*{Limitations}
Although the architecture of Dataverse can support multi-modal data such as image or video as well text data, the current implementation of Dataverse has yet to bring such features.
However, we plan to incorporate image and video support in future releases to maintain alignment with emerging research trends and evolving demands.

The Spark-based architecture of Dataverse necessitates tuning and optimization by experienced data engineers to achieve peak performance and scalability. While we have implemented reasonable defaults, we acknowledge that the current version may not fully unlock the potential inherent in Spark. 
For further optimization, we plan to add an automatic configuration feature that reasonably maximizes the Spark performance.

\section*{Ethics Statement}
We recognize that LLMs can reflect biases present in their training data, potentially generating skewed results across dimensions like race, gender, and age. While Dataverse incorporates bias mitigation techniques, ongoing monitoring and improvement are necessary.

The collection of massive datasets also raises privacy and copyright concerns. Dataverse aims to minimize these risks through anonymization and filtering, but vigilance is still required throughout the data acquisition and processing pipelines.

We are keenly aware of these ethical challenges in developing Dataverse. We are committed to continually enhancing Dataverse's capabilities to address bias, privacy, and potential misuse concerns. Our goal is to provide a powerful tool for advancing language AI while upholding robust ethical principles and mitigating societal risks to the greatest extent possible.

\bibliography{anthology, custom}

\clearpage
\appendix
\onecolumn

\section{Discussion}\label{sec:appn2}
\paragraph{Slow Progress in Open-Source Development for LLM Data Processing}
Notwithstanding the emergence of LLMs in the recent past, a paucity of widely adopted open-source solutions persists within the domain of data processing for these models. The substantial computational costs and significant computing power requirements have predominantly confined advancements to well-funded or large-scale organizations. Consequently, this has rendered LLM research inaccessible for individuals and smaller entities. Initially, the demand for open-source solutions in LLM research was not substantial. However, as smaller LLMs began to emerge, empowering individuals and smaller entities to participate in LLM research, the necessity for open-source solutions became increasingly evident. It is becoming progressively clear that to foster wider accessibility to LLM research and to facilitate greater participation in the field, there is an urgent need to accelerate the development of open-source solutions, with a particular emphasis on data processing.

\paragraph{Quintessential Elements for Developing Open-Source LLM Data Processing Solutions}
The quintessential elements for successfully developing open-source LLM data processing solutions can be distilled down to three key aspects: a user-friendly interface, cost-efficiency in data processing, and an automatic resource assessment tool.

Firstly, and most crucially, a user-friendly interface is imperative to ensure widespread community adoption. The objective should be to create an interface that is as intuitive as a computer's power button, thereby encouraging heightened user interaction and utilization.

However, a focus on user experience becomes effective only when underpinned by reliable, highly optimized, and cost-effective data processing capabilities. The utilization of data processing tools in the LLM era can be financially onerous. Consequently, these tools necessitate judicious calibration to ensure cost-effectiveness. Users must not encounter repeated attempts due to errors, as this not only adversely impacts the user experience but also exacerbates the associated costs.

Finally, the establishment of a system that automatically assesses the available computing power and evaluates its suitability for the given data is paramount. By doing so, it aims to preclude users from experiencing wasted time and frustration due to insufficient processing capabilities.

\paragraph{Registry-based System for Efficient Data Processor Management}
One might question the rationale for employing a registry, given that it can introduce complexity, particularly in multi-user systems due to synchronization issues. However, Dataverse operates as a single-user system, thereby eliminating the need for synchronizing the registry among users. This approach resolves the synchronization challenge and confers two key advantages. Firstly, it eliminates the necessity for importing data processor functions using relative paths, thereby simplifying the development process. Secondly, it enables an auto-registration system, alleviating the burden on users of having to manually save the data processing functions within the package. Instead, users are afforded the flexibility to implement their data processing functions in their preferred locations, without being constrained to a specific directory. Consequently, custom functions can be located in various environments, including Jupyter Notebooks, and can be seamlessly integrated into the ETL pipeline.

\paragraph{Block-Based Coding for Enhancing Experimentation}
The block-based coding approach essentially presents a data processor as a singular block, and an ETL Pipeline as a composite structure of multiple blocks. This design paradigm affords users the flexibility to add, remove, or reshuffle blocks with ease, achievable merely through configuration settings. Consequently, it enables users to effortlessly experiment with limitless combinations without the need to modify the codebase.

\paragraph{Batch Processing: An Enduring Approach in the Context of Large-Scale Data}
In the context of large-scale data, accuracy takes precedence over speed. The objective is not merely to thoughtlessly utilize the influx of data but rather to focus on producing high-quality and reliable data. To attain this, global assessment must encompass deduplication and ensuring a balanced perspective to avoid bias. This becomes substantially challenging in real-time data processing. As a result, Dataverse still heavily relies on batch processing, as it is designed for LLM data preparation, where accuracy and quality are paramount.

\paragraph{Naming Convention: Rationale Behind the Unconventional \_\_\_ Selection}
Naming large quantities of data processors, as is the case in Dataverse, presents two primary challenges: maintaining uniqueness and ensuring usability. With the potential for adding up to 10,000 data processors, guaranteeing unique identification can be daunting. Hence, the idea of categorization on two levels emerged, which not only ensures uniqueness but also makes the data processors easily identifiable and comprehensible.

However, a discussion emerged regarding whether or not to integrate these categories into the data processor's name. The confusion arises when functions like \textbf{remove} appear in multiple categories such as \textbf{deduplication} and \textbf{cleaning}. How do we clearly identify the difference? This problem was mitigated by asking users to provide the category and name as separate arguments within configurations. However, this proved to be cumbersome, and hence these elements were combined into a single character string. The separator underscores (\_\_\_) were then introduced to distinctively separate the ETL Category, ETL Sub-Category, and ETL Name. Consequently, the unconventional naming convention of [ETL Category]\_\_\_[ETL Sub-Category]\_\_\_[ETL Name] was employed.

\end{document}